\documentclass{article}
\pdfoutput=1

\usepackage{graphicx} 
\usepackage{CJKutf8}
\usepackage{url}
\usepackage[colorlinks=true, allcolors=blue]{hyperref}
\usepackage{amsmath}
\usepackage{amssymb}
\usepackage{amsthm}

\usepackage[letterpaper,top=2cm,bottom=2cm,left=3.5cm,right=3.5cm,marginparwidth=1.75cm]{geometry}
\usepackage{multicol}
\usepackage[ruled,linesnumbered]{algorithm2e}
\usepackage{algorithmic}
\usepackage{subfig}

\title{Early Warning Prediction with Automatic Labeling in Epilepsy Patients}

\author{
\normalsize{
Peng Zhang$^{1,2,3,}$\footnotemark[2]\ ,
Ting Gao$^{1,2,3,}$\footnotemark[1]\ ,
Jin Guo$^{1,2,3,}$\footnotemark[3]\ ,
Jinqiao Duan$^{4,5,3,}$\footnotemark[4]\ ,
Sergey Nikolenko$^{6,}$\footnotemark[5]
}\\[10pt]
\footnotesize{$^1$ School of Mathematics and Statistics, Huazhong University of Science and Technology, China} \\
\footnotesize{$^2$ Center for Mathematical Science, Huazhong University of Science and Technology, China} \\
\footnotesize{$^3$ Steklov-Wuhan Institute for Mathematical Exploration, Huazhong University of Science and Technology, China} \\
\footnotesize{$^4$ Department of Mathematics and Department of Physics, Great Bay University, China} \\
\footnotesize{$^5$ Dongguan Key Laboratory for Data Science and Intelligent Medicine, China}\\
\footnotesize{$^6$ St. Petersburg Department of the Steklov Mathematical Institute, Russia}}

\providecommand{\keywords}[1]{\textbf{\textit{Keywords:}} #1}
\date{}
\begin{document}
\begin{CJK}{UTF8}{gbsn}
\maketitle

\begin{abstract}
Early warning for epilepsy patients is crucial for their safety and well-being, in particular to prevent or minimize the severity of seizures. Through the patients' EEG data, we propose a meta learning framework to improve the prediction of early ictal signals. The proposed bi-level optimization framework can help automatically label noisy data at the early ictal stage, as well as optimize the training accuracy of the backbone model. To validate our approach, we conduct a series of experiments to predict seizure onset in various long-term windows, with LSTM and ResNet implemented as the baseline models. Our study demonstrates that not only the ictal prediction accuracy obtained by meta learning is significantly improved, but also the resulting model captures some intrinsic patterns of the noisy data that a single backbone model could not learn. As a result, the predicted probability generated by the meta network serves as a highly effective early warning indicator. 
\end{abstract}


\section{Introduction}
Nearly one third of the patients with epilepsy continue to have seizures despite optimal medication management \cite{ramgopal2014seizure}. According to the World Health Organization (WHO) \cite{WHO}, epilepsy affects approximately 50 million individuals worldwide, rendering it one of the most prevalent neurological disorders on a global scale. Seizures result from abnormal discharges originating in a cluster of brain cells \cite{gentiletti2017changes, cogan2017multi}. These discharges can originate in various regions of the brain, leading to a spectrum of seizure manifestations, ranging from momentary lapses of consciousness or muscle reflexes to prolonged and intense convulsive episodes.

\footnotetext[2]{Email: \texttt{kazusa\_zp@hust.edu.cn}}
\footnotetext[1]{Email:  \texttt{tgao0716@hust.edu.cn}}
\footnotetext[3]{Email: \texttt{jinguo0805@hust.edu.cn}}
\footnotetext[4]{Email: \texttt{duan@gbu.edu.cn}}
\footnotetext[5]{Email: \texttt{sergey@logic.pdmi.ras.ru}}

Individuals with epilepsy often experience a heightened prevalence of physical issues such as seizure-related fractures and abrasions \cite{thurman2017burden}. Additionally, a greater proportion of individuals with epilepsy also grapple with psychological disorders, including anxiety and depression \cite{schabert2022incidence}. Furthermore, there exists substantial evidence indicating that individuals with uncontrolled epilepsy over an extended period face an elevated risk of enduring memory impairment, depression, anxiety, suicide, and other psychiatric conditions \cite{ott2003behavioral}. Despite the existence of certain clinical treatments, a longitudinal study spanning over 20 years on idiopathic generalised epilepsies (IGEs) revealed that individuals with epilepsy still report a diminished quality of life \cite{camfield2010idiopathic}, primarily attributed to the unpredictability of seizures and their adverse consequences . Consequently, the development of early warning systems for epilepsy holds paramount significance in addressing these challenges.

Electroencephalography (EEG) serves as a highly effective diagnostic instrument for investigating the functional intricacies of the brain during epileptic seizures. Extensive research has been dedicated to the prediction and treatment of epilepsy through the utilization of EEG \cite{adeli2003analysis,yang2019strategy,birjandtalab2017automated,cao2017epileptic,sharan2020epileptic,san2019classification,hussein2019optimized}. EEG signals, characterized by their non-Gaussian and non-smooth properties, are instrumental in quantifying electrical brain activity.  This, in turn, aids in diagnosing various types of brain disorders, so the analysis of EEG measurements plays a key role in distinguishing between normal and aberrant brain function.
Throughout the past century, researchers have grappled with the challenges inherent in epilepsy detection and prediction. Given that EEG signals represent a key resource for monitoring brain activity preceding, during, and following epileptic seizures, epileptic seizure prediction research has primarily centered on the analysis of EEG recordings. 

Deep learning architectures have found applications across various medical domains, including clinical imaging \cite{lee2017deep}, genomics and proteomics \cite{asgari2015continuous}, and disease prediction \cite{khan2017focal}. Notably, deep learning algorithms have demonstrated their efficacy in discerning intricate patterns within high-dimensional data, particularly in the context of EEG data classification~\cite{Roy_2019,10.1007/978-3-031-06427-2_28,KIM2023120054}. In particular, 
to introduce a universally applicable method for all patients Truong et al.~\cite{truong2017generalised} propose a prediction approach based on convolutional neural networks (CNNs). Their methodology involved the transformation of raw EEG data into a 2-dimensional matrix through a short-term Fourier transform (STFT). Subsequently, this image is input into a CNN to facilitate feature extraction and classification of pre-ictal and inter-ictal states. Generally speaking, CNN-based architectures remain standard for EEG processing, in particular for epileptic seizure detection, to this day~\cite{10.1007/s11042-023-15052-2,GG22,10.3389/fneur.2020.00375}.

It is important to note that the vast majority of high-performance seizure prediction algorithms operate in a fully supervised setting, training exclusively on labeled data.  Nevertheless, it is essential to recognize that the process of labeling seizure data is manual and costly, it requires the expertise of a neurologist and consumes substantial resources. To address this challenge, Truong et al.~\cite{truong2019epileptic} employ a Generative Adversarial Network (GAN) for unsupervised training. They feed STFT spectrograms of EEG data into the GAN and utilize trained discriminators as features for seizure prediction. This unsupervised training approach is particularly valuable as it not only enables real-time predictions using EEG recordings but also eliminates the need for manual feature extraction. In a different approach, Tsiouris et al.~\cite{hinton2011machine} employ long short-term memory (LSTM) recurrent networks for seizure prediction. They conduct a comparative assessment of various LSTM architectures using randomly selected input segment sizes ranging from 5 to 50 seconds. Their evaluation encompasses three distinct LSTM architectures, each utilizing feature vectors derived from EEG segments as input. These feature vectors contain a diverse array of attributes, incorporating temporal and frequency domain characteristics, as well as local and global metrics derived from graph theory. LSTMs, and RNNs in general, are another common way to process EEG data due to its sequential nature~\cite{diagnostics13040773}, and generative adversarial networks have been used many times for synthetic data generation~\cite{N21}, including synthetic EEG generation~\cite{PMID:37038142,Hartmann2018EEGGANGA}.

While the above research has demonstrated effectiveness in predicting seizures, applications of deep learning for early seizure warning still encounters several challenges since EEG data has high frequency, high dimension, and comes under the influence of complex noise distributions. 


The first challenge revolves around the high cost of accurate and reliable labels. Precise labeling of epileptic seizure and non-seizure EEG data is of fundamental importance for model training, but the process of labeling EEG data necessitates the expertise of physicians and can be a time- and resource-intensive endeavor. For example, a recent work~\cite{GG22} presents a new CNN-based architecture based on a dataset of only 79 EEG recordings labeled by three experts. Furthermore, labels produced by physicians may not always be correct. EEG data labeling by physicians often relies on their individual experiences, and it is not uncommon for different physicians to assign different labels to the same dataset.

The second challenge is that the prediction for the onset of seizures under ambiguous states between resting and seizure states is difficult, especially for long-term prediction tasks.
Additionally, epileptic seizures can be regarded as sudden and unfrequent transitions in brain dynamics, making it challenging to capture these rare events in time.

In this work, we present a new framework tackling the above issues to make early warnings for epileptic seizures. In response to the first and second challenges, we have implemented enhancements to the method of meta label correction \cite{zheng2021meta}. This refinement enables the training of a primary model capable of providing highly accurate predicted labels even when supplied with a limited quantity of clean data. As for the third challenge, we develop the capability to predict early warning signals of critical transitions via tipping point detection.

To summarize, we develop a new method to detect the early warning signals integrating both data-driven models and the tipping dynamics. Our key contributions in this work are as follows:
\begin{itemize}
\item \textbf{meta labeling}: we split the original data into a clean part and a noisy part and subsequently integrate meta-learning into conventional methods in order to optimize both labeled information from clean data and the complex feature patterns from noisy data with automatically generated labels; 

\item \textbf{real data experiments}: our experiments conducted on real-world EEG data obtained from epilepsy patients reveal a substantial enhancement in predicted accuracy; we additionally report a series of experiments conducted with various input and output window sizes;

\item \textbf{tipping point indicator}: we find that the \textit{predicted probability} obtained by the meta learning framework can serve as an effective early warning indicator for tipping phenomena; our study illustrates that meta learning can capture some patterns between rest states and seizure states that a single backbone model could not. 
\end{itemize}

\section{Meta labeling and tipping phenomena}

\subsection{Meta Label Correction}












High-quality labels are crucial for enhancing the training performance of neural networks. Following the assumptions of~\cite{charikar2017learning,veit2017learning}, we assume that labels with noise can be categorized into two distinct groups: one dataset with clean or trusted labels and other datasets with noisy or weak labels. In many cases, the dataset with clean labels is smaller than those with noisy labels, primarily due to the limited availability of clean labels and the associated high cost of labeling. Direct training on these smaller clean datasets can result in suboptimal performance, with a pronounced risk of overfitting. Similarly, training exclusively on noisy datasets---or a mix of both clean and noisy datasets---may also yield unsatisfactory outcomes, as large-scale models can inadvertently fit and memorize the noise~\cite{zhang2021understanding}.

Our objective is to identify precursor features that contribute to seizures, enabling us to distinguish data oscillations caused by other behaviors. For epileptic patients, seizures often last only 30 seconds to 2 minutes, which makes seizure data scarce compared to non-seizure data and leads to data imbalance issues. Moreover, accurate labeling of seizure and non-seizure data serves as the foundation for model training, a process typically requiring the expertise of a physician and demanding substantial time and effort. Yang et al. propose the MMLC method to handle financial data with partial feature information and poor label quality~\cite{luxuan2023MultitaskMeta}. Here, we have far more clean data than noisy data. Our dataset is partitioned into two subsets: a smaller dataset with potentially noisy or no labels during the period when seizures are possible and a larger dataset with clean and reliable labels during non-seizure period or post-ictal period when the seizures have concluded.

Our aim is to employ clean data to facilitate the labeling of noisy data. Unlike conventional label generation methods, our approach relies on making label judgments for the data based on their predicted time domain, obviating the need to directly obtain labels for the noisy data. This concept operates with a two-layer optimization paradigm. Initially, we employ a meta model to predict labels for the noisy data and subsequently refine our meta model using clean data. The main network utilizes these corrected labels to classify noisy data.



\subsection{Early Warning Indicators}
Under long-term noise disturbances, a dynamical system may experience enormous fluctuations, which could lead to transitions between metastable states \cite{gao2016dynamical,guo2023deep}.
Sudden transitions between alternative states are a ubiquitous phenomenon within complex systems, spanning a wide spectrum of domains ranging from cellular regulatory networks to neural processes and ecological systems \cite{scheffer2009early,ashwin2015pattern}. Numerous instances of abrupt regime shifts have been observed in diverse research fields, including collapses of ecosystems \cite{hirota2011global,wang2012flickering}, abrupt climate shifts \cite{lenton2012early,drijfhout2015catalogue}, and onset of certain disease states such as atrial fibrillation \cite{quail2015predicting} or epileptic seizures \cite{meisel2012scaling}. Warning signals for impending critical transitions are highly desirable, owing to the formidable challenges associated with restoring a system to its prior state subsequent to the occurrence of such transitions \cite{folke2004regime,scheffer2001catastrophic}. Lade and Gross \cite{lade2012early} introduce a method for the identification of early warning signals, offering a comprehensive integration of diverse information sources and data within the context of a generalized model framework. Proverbio et al. \cite{proverbio2023systematic} undertake a systematic exploration of the characteristics and effectiveness of dynamical early warning signals across various conditions of system deterioration.

An increase in the variance within the pattern of fluctuations \cite{carpenter2006rising} represents a plausible outcome of critical slowing down when a system approaches a critical transition. This phenomenon can be rigorously demonstrated through formal mathematical representations \cite{biggs2009turning} and intuitively grasped.  As the eigenvalue of the system approaches zero, the influence of perturbations becomes increasingly enduring, resulting in an amplified accumulation effect that subsequently augments the variance of the system's state variable.  In principle, one might anticipate that critical slowing down could impair the system's capacity to trace these fluctuations, potentially leading to a opposite effect on the variance \cite{berglund2002metastability,berglund2006noise}.  Nevertheless, investigations into mathematical models consistently reveal that an elevation in variance typically emerges and becomes discernible well in advance of the occurrence of a critical transition \cite{carpenter2006rising}. Neubert and Caswell \cite{aldasoro2018early} introduce a pair of quantitative metrics for characterizing transient dynamics, namely \textit{reactivity} and the \textit{amplification envelope}. These metrics can be computed by utilizing the Jacobian matrix in the vicinity of a stable equilibrium. \textit{Reactivity} offers an estimation of the system's susceptibility to immediate perturbation-induced growth, while the maximum \textit{amplification} quantifies the highest magnitude that a perturbation can transiently achieve relative to the system's equilibrium.

In this work, one of our goals is to extract fluctuation patterns from data via the meta learning network as an early warning indicator for state transitions in EEG time series signals.

\section{Our Method}
\subsection{Automatic Label Optimization through Meta Learning}

\def\balpha{\boldsymbol{\alpha}}
\def\bomega{\boldsymbol{\omega}}

The meta-model comprises two neural networks: a \textit{meta network} and a \textit{main network}. The \textit{meta network}'s objective is to enhance label accuracy for noisy data, while the \textit{main network}'s role is to achieve correct classification. We denote the parameters of the \textit{meta network}, considered as meta-knowledge, as $\balpha$, denoting the meta network as $g_{\balpha}(X', Y')$. Conversely, the \textit{main network}'s parameters are denoted as $\bomega$, and it computes the function $f_{\bomega}(X', y_c)$, where $y_c$ represents the outcome obtained from the \textit{meta network}. 

 \begin{figure}[!t]
     \centering
     \includegraphics[width=1.0\linewidth]{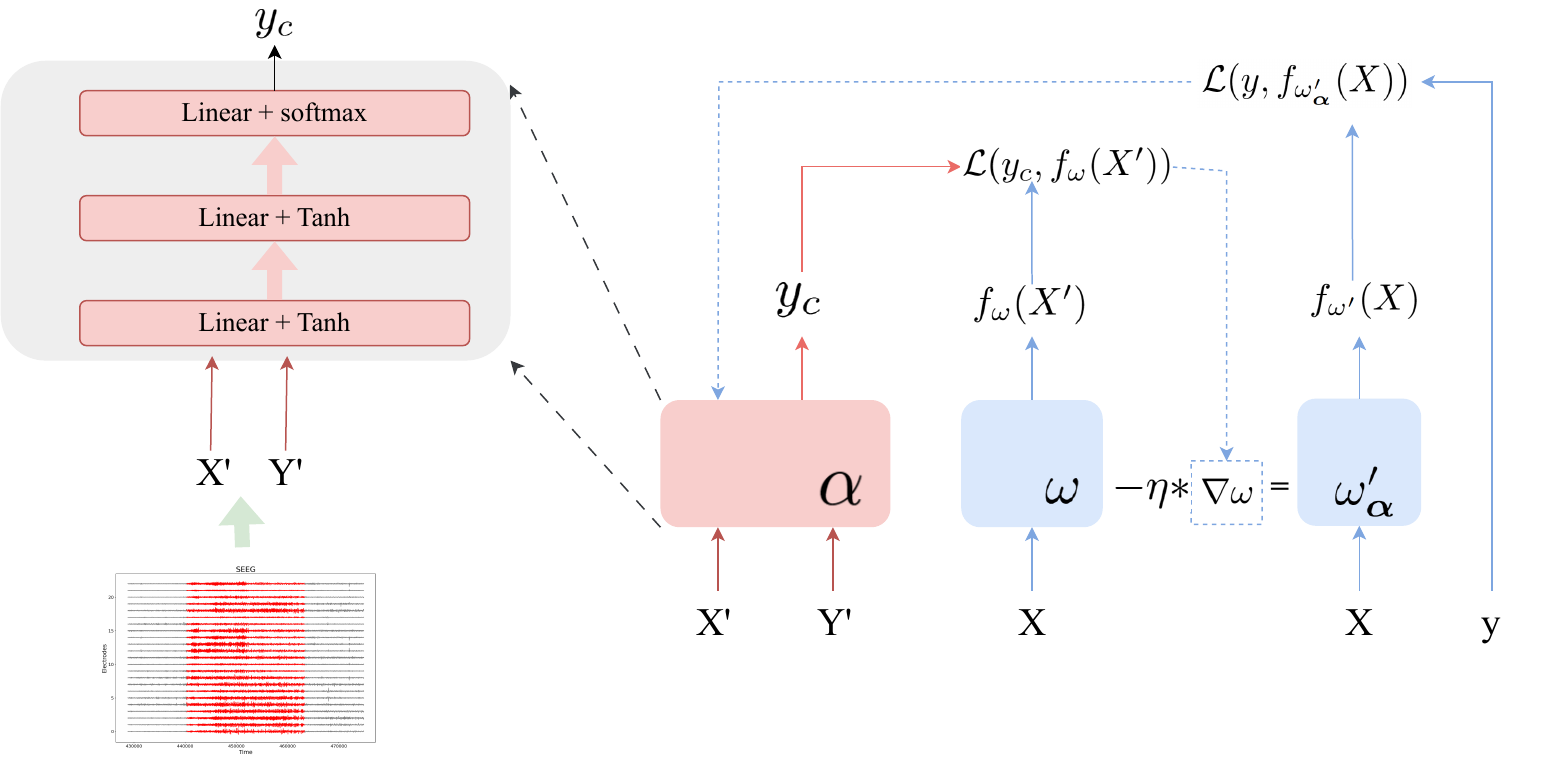}
     \caption{The framework of meta label generation model.}
     \label{model}
 \end{figure}

The training process is illustrated in Figure~\ref{model}. The data we feed into the meta network is divided into two parts. One part is historical data $X'$ input as a time window of length $h\in \mathbb{R}^+$, that is, $X'$ represents the EEG time series from time $t$ to $t+h$. The other part is the known data $Y'$, and its time window size is denoted as $m\in \mathbb{R}^+$, that is, $Y'$ represents the EEG time series from time $t+h$ to time $t+h+m$; both $X'$ and $Y'$ are sequential in time. The meta network learns the label $y_c$ of data $Y'$, where $y_c\in \{0,1\}$. If the pattern of $Y'$ indicates that the patient had seizures during this time period, the meta network should output $y_c=1$, otherwise it should output $y_c=0$. As for the main network, its objective is to learn the label of the next (unknown) EEG signal window $Y$ of the current time series data $X$, that is, the main network does not know the sequential data $Y$ and is supposed to predict the label of unknown $Y$ with known data $X$. This approach enables us to extract additional information about the prediction horizon and historical data without being influenced by noisy labels. We have already mentioned that labels given manually by experts can be subjective and are certainly expensive; our approach can avoid this problem by generating noisy data labels.

In every batch, we obtain both noisy and clean sample data. We initially input the noisy samples, along with their corresponding $(X', Y')$ pairs, into the \textit{meta network} to derive accurate labels. Subsequently, the noisy data $X'$ is utilized as input for the current task's classifier to generate the corresponding predicted labels. We update the \textit{main network} classifier's parameters based on the loss between the corrected and predicted labels. Following this, the clean data pair $(X, y)$ is introduced to the new classifier, and the classification loss is computed to facilitate the \textit{meta network}'s update. 

The bilevel optimization of our model can be formulated as follows:
\begin{align}
    \min_{\balpha}\ \  & \mathbb{E}_{(X,y)\in D}\mathcal{L}_D\left(f_{\bomega^*_{\balpha}}(X),y\right),\text{ where}\\
\bomega^*_{\balpha} &=\arg\,\min\limits_{\bomega}\mathbb{E}_{(X',Y')\in D'}\left[\mathcal{L}_{D'}\left(f_{\bomega}(X'),g_{\balpha}(X',Y')\right)\right],
\end{align}
and where the loss function is defined as $$\mathcal{L}(a,b)=\frac{1}{N}\sum_{i=1}^N (b_i\ln{(a_i)}+(1-b_i)\ln{(1-a_i)}),$$ 
i.e., the cross entropy loss for $a, b\in \mathbb{R}^{N\times 2}$. Our algorithm is shown in Algorithm~\ref{algorithm}.

\begin{algorithm}[!t]
\caption{Meta learning for noise label generation}\label{algorithm}
\KwIn{Clean dataset $D=\{X,y\}$, noisy dataset $D'=\{X',Y'\}$, learning rates $\mu,\eta$}
    \emph{Initialize main network weights $\boldsymbol{\omega}$ and meta network weights $\alpha$}\;
    
    \For{$n\leftarrow 0$ \KwTo{episodes}}
    {   
    \For{$k\leftarrow 0$ \KwTo{K}}{
     \emph{$\mathcal{L}_{D'}=\frac{1}{N}\sum_{i=1}^N(g_\alpha(X',Y')\ln{(f_{\bomega}(X')+(1-g_\alpha(X',Y')\ln{(1-f_{\bomega(X')}}}))$}\;
     
     \emph{Update main network parameters by gradient descent:} $\bomega\leftarrow w-\mu \nabla_{\bomega}\mathcal{L}_{D'}.$}
     
     \emph{$\mathcal{L}_D=\frac{1}{N}\sum_{i=1}^N(y\ln{(f_{\bomega}(X)+(1-y\ln{(1-f_{\bomega(X)}}}))$}\;
     \emph{Update meta network parameters by gradient descent: $\balpha\leftarrow \balpha-\eta \nabla_{\balpha}\mathcal{L}_{D}$}

    }
\end{algorithm}

Updating the meta model presents a formidable challenge, as the resulting loss function does not explicitly feature $\alpha$. Consequently, we have to compute $\frac{\partial \bomega^*_{\boldsymbol{\alpha}}}{\partial \boldsymbol{\alpha}}$, which involves second-order gradient computations. Several different methods exist for this purpose. To circumvent resource-intensive demands, we employ the Taylor expansion method, as detailed in \cite{zheng2021meta}, to compute the Hessian matrix $\frac{\partial^2}{\partial \bomega^2} \mathcal{L}_{D^{\prime}}\left(\boldsymbol{\alpha}, 
 \bomega\right)$.



Denote the meta loss as $\mathcal{L}_{D}\left(\bomega^*_{\boldsymbol{\alpha}}\right) \triangleq \mathbb{E}_{(\mathbf{X}, y) \in D} \ell\left(y, f_{\mathbf{w}^*_{\boldsymbol{\alpha}}}(\mathbf{X})\right)$. Then we have 
\begin{align}
\min _{\boldsymbol{\alpha}} \mathcal{L}_{D}\left(\mathbf{w}^*_{\boldsymbol{\alpha}}\right) & \approx \mathcal{L}_{D}\left(\mathbf{w}^{\prime}_{\boldsymbol{\alpha}}\right) \\
& =\mathcal{L}_{D}\left(\bomega-\eta \nabla_{\bomega} \mathcal{L}_{D^{\prime}}\left(\boldsymbol{\alpha}, \bomega\right)\right).
\end{align}
Subsequently, the update rule for the meta-knowledge $\boldsymbol{\alpha}$ can be written as follows:
\begin{align}
\boldsymbol{\alpha}=\boldsymbol{\alpha}-\mu \nabla_{\boldsymbol{\alpha}}  \mathcal{L}_{D}\left(\bomega^{\prime}_{\boldsymbol{\alpha}}\right).
\end{align}
Then, the meta-parameter gradient from previous $T$ step  is shown as:
\begin{multline}
\frac{\partial \mathcal{L}_{D}\left(\bomega^{\prime}_{\boldsymbol{\alpha}}\right)}{\partial \alpha}= g_{\bomega^{\prime}}\left(I-\eta \nabla_{\bomega, \bomega} \mathcal{L}_{D^{\prime}}\left(\boldsymbol{\alpha}, \bomega\right)\right) \frac{g_{\bomega^{\top}}}{\left\|g_{\bomega}\right\|^2} \frac{\partial \mathcal{L}_{D}\left(\bomega\right)}{\partial \boldsymbol{\alpha}} - \\ -\eta \nabla_{\boldsymbol{\alpha}}\left(\nabla_{\bomega}^{\top} \mathcal{L}_{D^{\prime}}\left(\boldsymbol{\alpha}, \bomega\right) \nabla_{\bomega^{\prime}} \mathcal{L}_{D}\left(\bomega^{\prime}\right)\right).
\end{multline}

\subsection{Early Warning Probability Indicator}
The capacity to discern transitions between meta-stable states assumes a crucial role in the prediction and regulation of brain function. In our model, the objective of the main network is to learn the label of the next sequential data $Y$ (unknown) based on the current time series data $X$, that is, the main network does not know the sequential data $Y$ and should predict the label of unknown $Y$ with known data $X$. Namely, the output of the main network is the probability of epileptic seizures occurring within the subsequent sequential time period following $[t,t+h]$. 

An escalation in the variance within the pattern of fluctuations, as established in the work~\cite{carpenter2006rising}, can serve as a credible indicator when a system is near a tipping point and is experiencing a critical slowing down. We adopt the value of the main network's output as an early warning indicator in the context of epilepsy, i.e.,
\begin{equation}I_{P}=f_{\bomega}(X),\end{equation}
which could be seen as the probability of ictal seizure occurring in the next time window. We conduct experiments in Section~\ref{comp} to compare our early warning indicator $I_{P}$ with the variance of time series data.

\section{Experiments}
\subsection{Data Processing}

\begin{figure}[!t]
\centering            
  \subfloat[ ]
  {\label{EEG data}
      \includegraphics[width=0.48\linewidth]{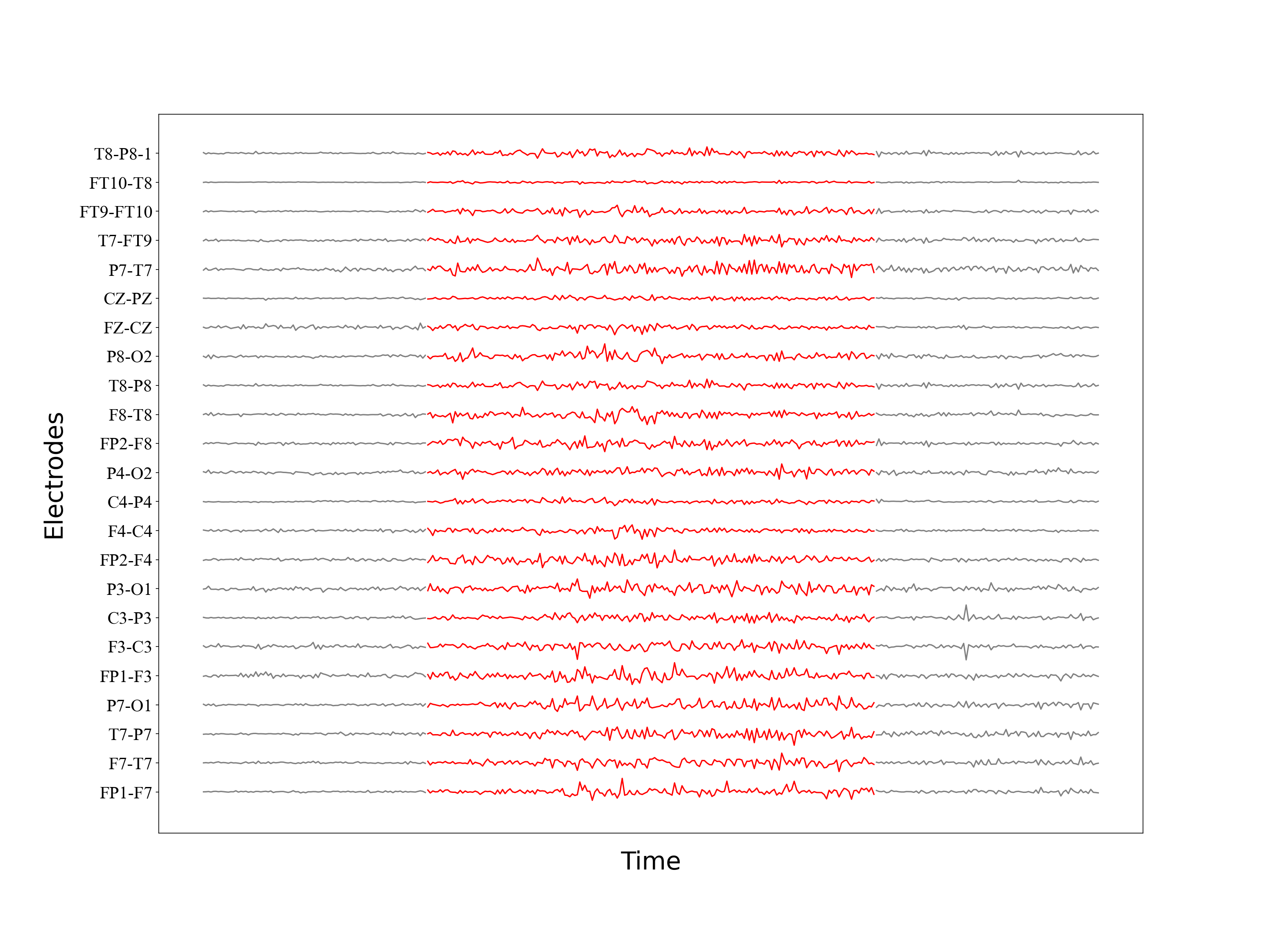}
  }
  \subfloat[ ]
  {\label{var}
    \includegraphics[width=0.48\linewidth]{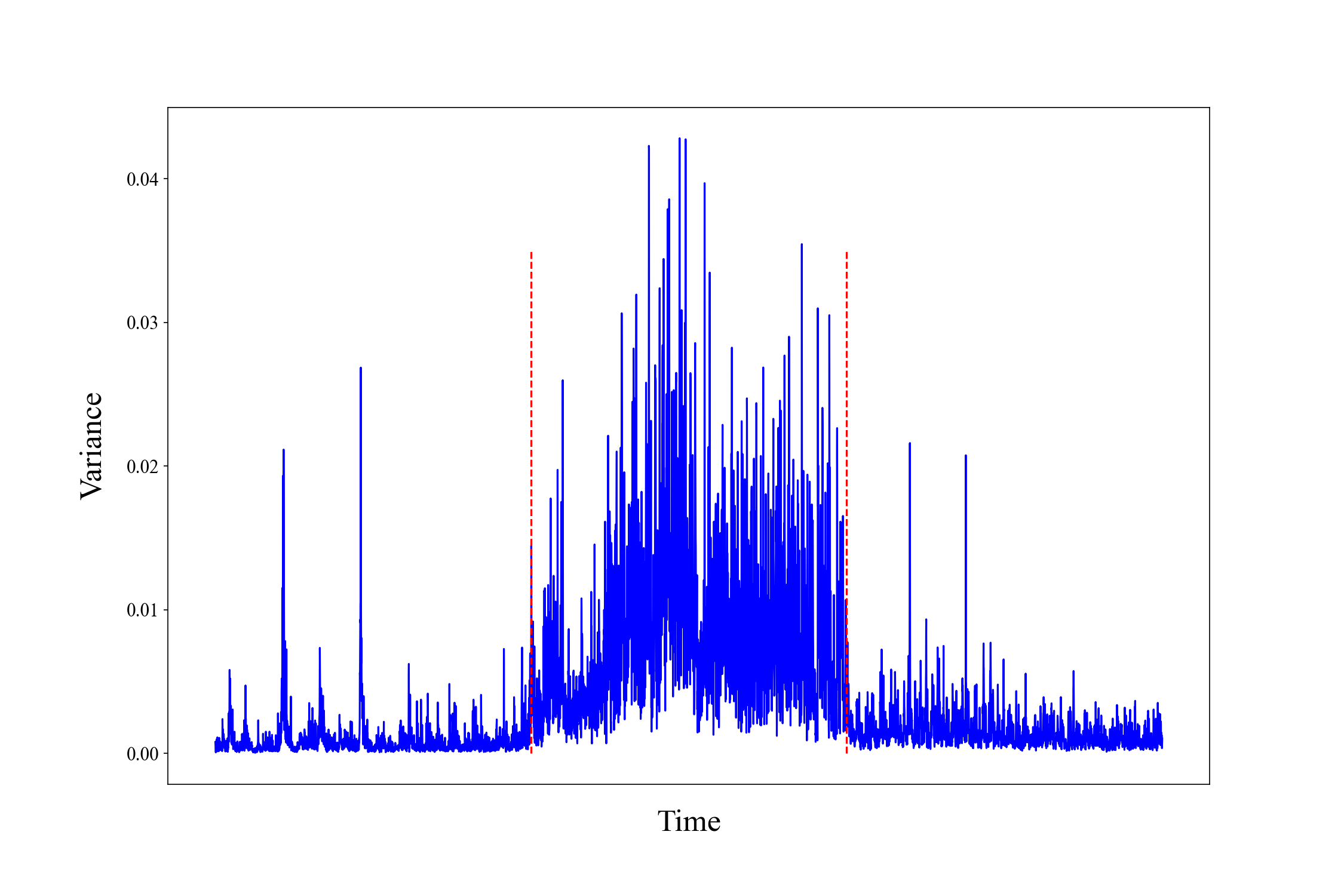}
  }
  
  \caption{A sample of our EEG data: (a) time series for the electrodes; the red part shows seizure states, and the gray part represents rest states; (b) variance of the EEG data.}    
\end{figure}

The Scalp EEG data is recorded via the bipolar montage method \cite{zaveri2006use}, where the voltage of each electrode is linked and compared with neighboring electrodes to form a chain of electrodes. Figure~\ref{EEG data} shows a sample of the data, and the red line represents time intervals identified as seizure occurrences by experienced physicians in this case. The horizontal axis represents the time of the test, while the vertical axis corresponds to specific brain regions where electrodes were inserted. Each segment of EEG data was recorded with 23 channel sampling rate of 256 Hz and recording duration of 1 hour, during which 5 seizures were recorded for the subject. Our dataset is derived from this EEG data, collected at a frequency of 32Hz. Due to the high cost of acquiring such data, the size of the sample under examination remains modest. Consequently, the development of an early warning system based upon this limited dataset becomes especially challenging. We collected data from \emph{five epileptic episodes} of an epileptic patient and used \emph{four epileptic episodes} to generate the \emph{training set} and \emph{the remaining one episode} as the \emph{test set}. Figure~\ref{EEG data} shows one epileptic episode. 


To establish the benchmark, we employ a sliding window technique, shifting the window left and right with a 0.5-second step. We set the time window of $X$ to encompass 10 seconds, 15 seconds, and 20 seconds, aligning precisely with durations of historical data. For $Y$, on the other hand, we have considered time windows spanning 5, 10, 15, and 20 seconds; these windows indicate the length of our prediction horizon. 

Our dataset is divided into clean data and noisy data. Noisy data refers to data instances that are especially challenging for labeling due to their inherent complexity and lack of clear, unequivocal categorization. To be precise, we define an indeterminate zone around the clinically established onset time of epilepsy as the demarcation point, i.e., we define a time window spanning $\pm 10$ seconds around this moment, amounting to 40 samples in every epileptic episode, as noisy data. Then we expand the time window both forward and backward, collecting 40 clean data samples with labels 0 and 1 respectively on every side. In instances where the right boundary of the $Y$-window intersects with this indeterminate region, we designate the corresponding window $X$ and $Y$ as noisy data; we denote noisy data as $X'$ and $Y'$ and obtain the resulting noisy dataset $D'=\{X',Y'\}$. Noisy data does not have fixed determined labels, and our objective is to assign accurate labels to this data category without relying on expert annotations. In essence, we aim to generate precise labels for this noisy dataset autonomously. The rest of the data, which falls outside the indeterminate zones around onset times, is categorized as clean data. The clean dataset is partitioned into two distinct segments: one segment corresponds to the data obtained during the resting state, with label $0$, and the other segment comprises data recorded during the epileptic state, with label $1$. The clean dataset is denoted as $D=\{X,y\}$, where $y\in \{0,1\}$ is the label. 



Different electrodes have data with different orders of magnitude, so all data was normalized, and then noisy and clean data were divided according to different tasks. Figure~\ref{var} shows a sample of the covariance of the EEG data time series, with the red dotted line showing an epileptic seizure in the sample. The figure illustrates how growing variance could be seen as an indicator for an upcoming abrupt transition, which is consistent with~\cite{carpenter2006rising}.


\subsection{Experimental Results}\label{comp}


 



\begin{table}[!t]
\centering
    \begin{tabular}{c|ccc|ccc|ccc}
    \hline
    Method &  LSTM & ResNet & \textbf{Ours} & LSTM & ResNet & \textbf{Ours} & LSTM & ResNet & \textbf{Ours}\\
    \hline
       $|Y|$  & \multicolumn{3}{c|}{$|X| = 10s$} & \multicolumn{3}{c|}{$|X| = 20s$} & \multicolumn{3}{c}{$|X| = 30s$} \\
    \cline{2-10}
        5s & 93.8\% & 92.5\% & 97.5\% &95.0\% & 95.0\%& 98.8\% & 96.3\% & 93.8\% & 97.5\% \\
        10s & 85.0\% &86.3\% & 92.5\% & 87.5\% & 86.3\%& 91.7\%& 90.0\% & 88.8\% & 94.4\% \\
        15s & 72.5\% &77.5\% & 83.8\% & 75.0\% & 78.8\%& 86.3\%& 76.3\% & 80.0\% & 88.9\% \\
        20s & 62.5\% & 71.3\% & 76.3\% & 63.8\% & 70.0\% & 78.8\% & 66.3\% & 73.8\% & 82.5\% \\    
        \hline
    \end{tabular}
    \caption{The accuracy of LSTM, ResNet, and our method. Columns correspond to the duration of the input data time $X$ (shown in row 2), and rows correspond to different lengths of the prediction window $Y$.}
    \label{accu}
\end{table}

\paragraph{Performance comparison.}  To test the performance of our approach, we have conducted experiments with a number of different models. Table~\ref{accu} provides a comprehensive overview of our classification results, comparing the performance of the baseline LSTM model, the ResNet model, and our proposed model. Here, we use LSTM for the meta network and ResNet for the main network. Columns in Table~\ref{accu} show the length of the input (historical) time window $|X|$, and the rows correspond to different sizes of the prediction horizon $|Y|$. It is clear that as the prediction horizon $|Y|$ (shown in the first column) grows, it becomes increasingly difficult to make accurate predictions for all three models. The accuracy of the LSTM model rises significantly as the length of the history duration becomes longer, because longer sequences in the input data contain more information. ResNet, on the other hand, has no particularly pronounced dependence on the input time length. Additionally, LSTM shows a very promising predictive ability over short periods of time, but the accuracy decreases rapidly as the prediction horizon length increases. The ResNet model, on the other hand, still shows a good degree of adaptation for prediction lengths greater than 10 seconds.

Throughout all settings, our model shows substantial improvements in prediction accuracy. Naturally, all three models achieve higher accuracy values at shorter prediction ranges, but our model demonstrates excellent results for longer noisy data as well. Our approach reduces the error rate by at least 2x in most cases: for 5s prediction horizons our model reduces the error rate from 5-7\% to 2-2.5\%, for 10s, from 10-15\% to 5-8\%, and for 15 and 20 seconds the error goes down from 20-30\% or higher to 15-25\%.
These results suggest that our approach based on meta-learning can very significantly improve the performance of baseline models in epileptic seizure prediction tasks.

\begin{figure}[!t]
\centering            
  \subfloat[LSTM, $|X|=20$, $|Y|=5$]{
      \label{lstm pre 5}\includegraphics[width=0.32\textwidth]{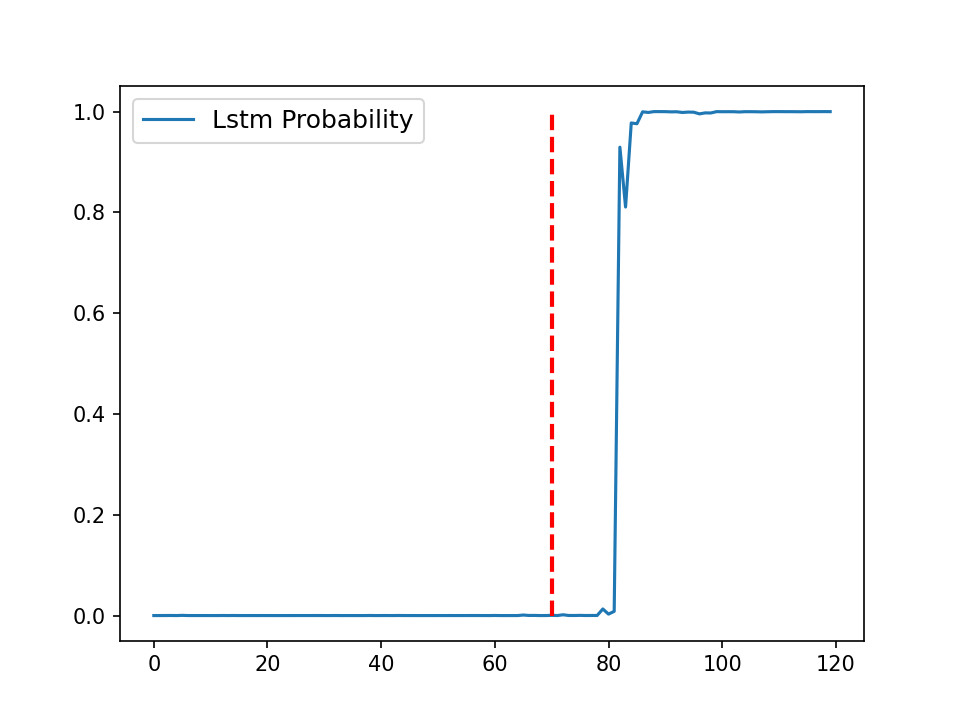}
      }
  \subfloat[ResNet, $|X|=20$, $|Y|=5$]
  {
      \label{renet pre 5}\includegraphics[width=0.32\textwidth]{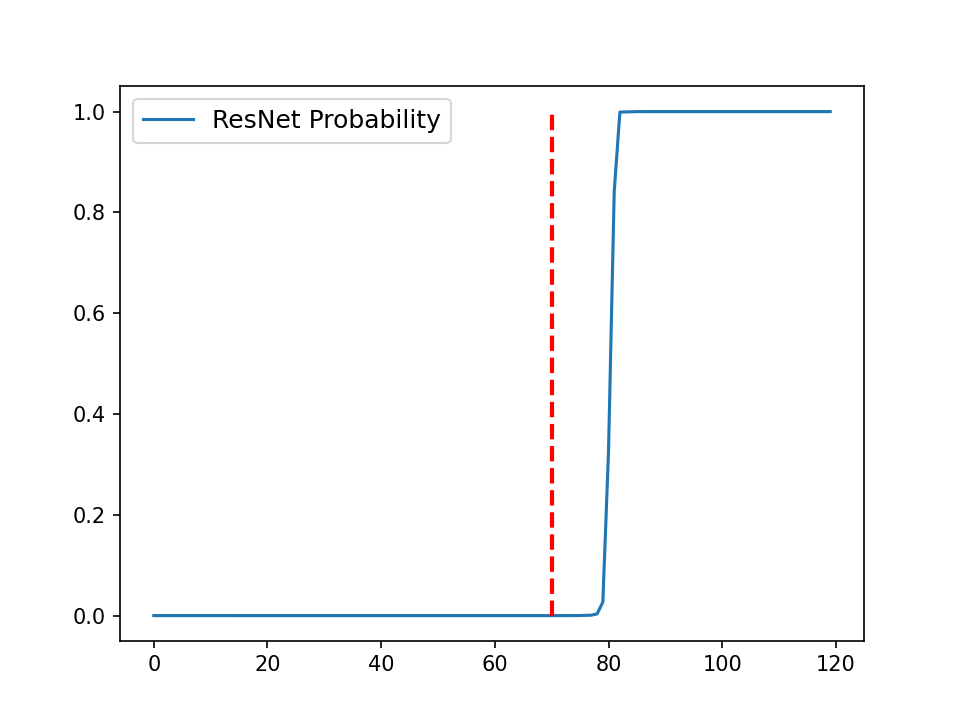}
  }
  \subfloat[Our model, $|X|=20$, $|Y|=5$]
  {
      \label{meta pre 5}\includegraphics[width=0.32\textwidth]{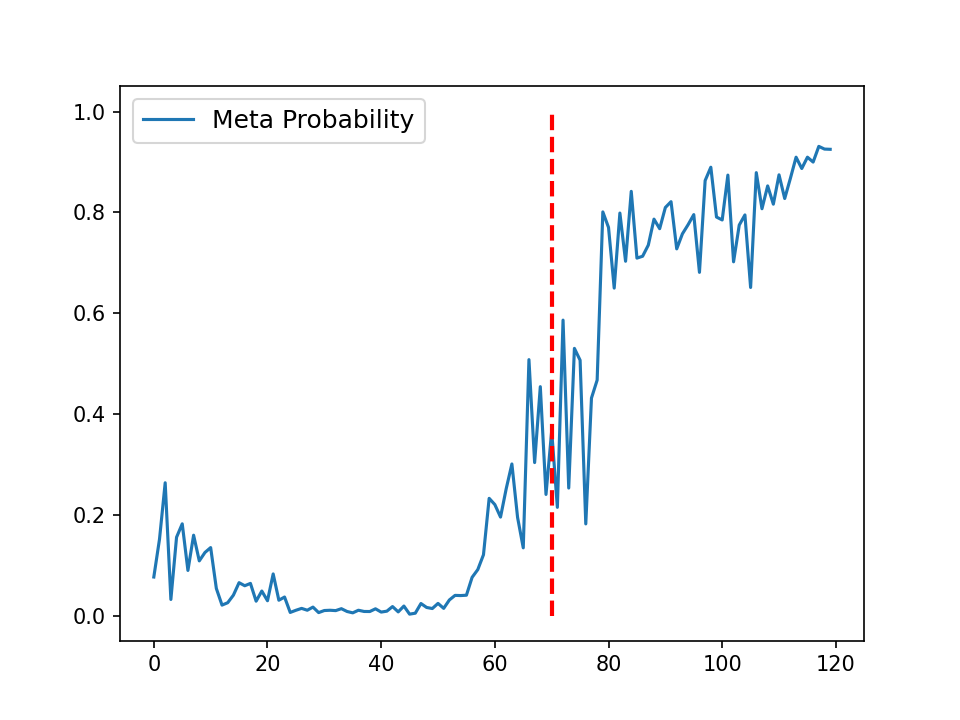}
  }

    \subfloat[LSTM, $|X|=20$, $|Y|=10$]{
      \label{lstm pre 10}\includegraphics[width=0.32\textwidth]{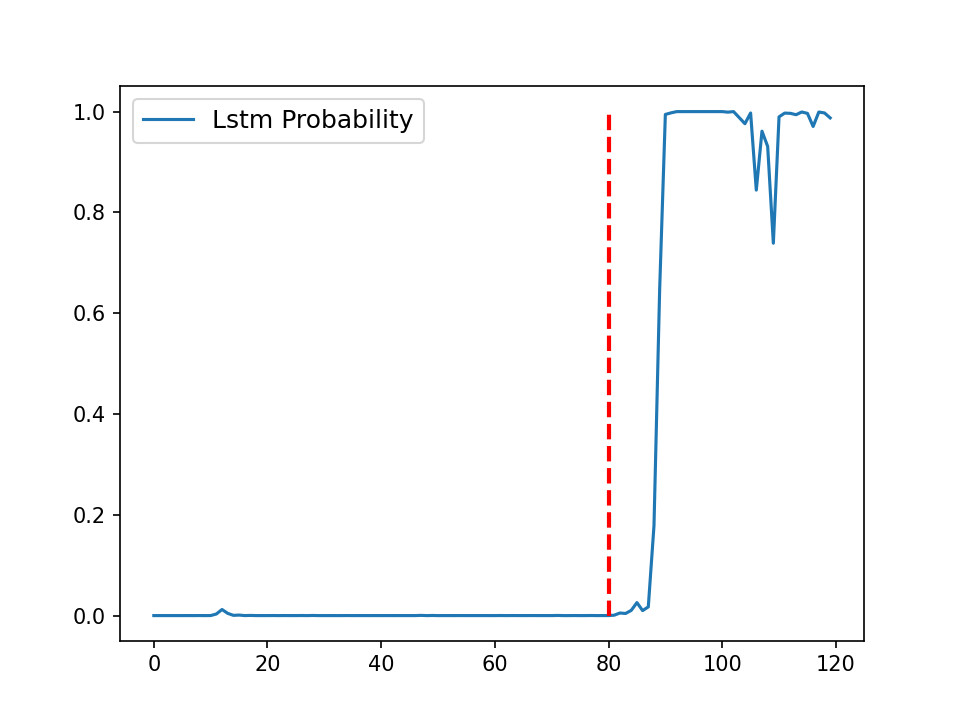}
      }
  \subfloat[ResNet, $|X|=20$, $|Y|=10$]
  {
      \label{renet pre 10}\includegraphics[width=0.32\textwidth]{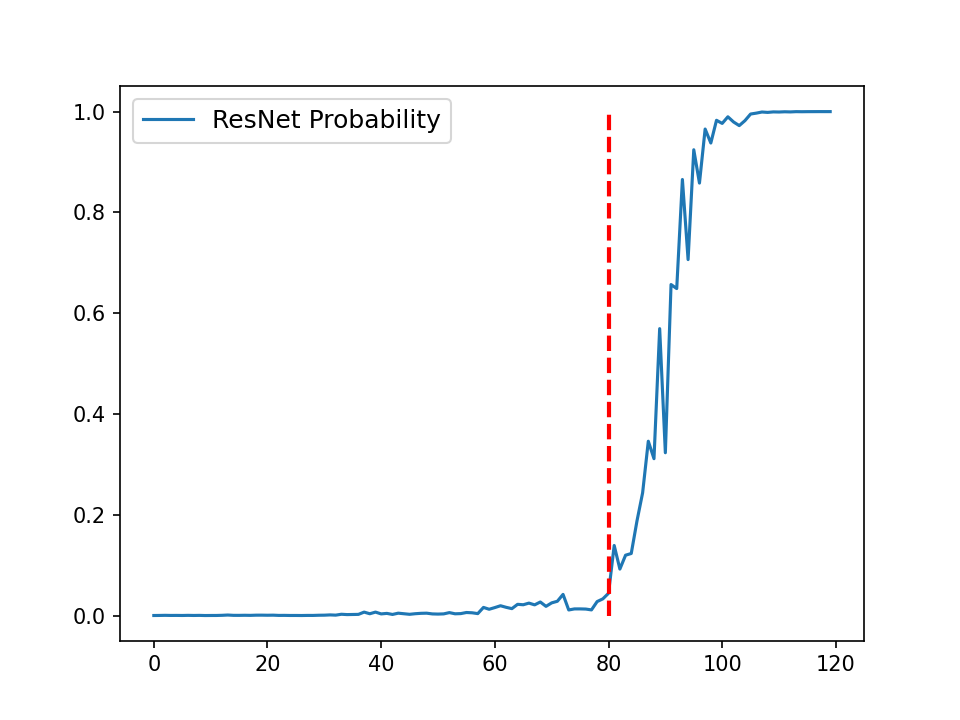}
  }
  \subfloat[Our model, $|X|=20$, $|Y|=10$]
  {
      \label{meta pre 10}\includegraphics[width=0.32\textwidth]{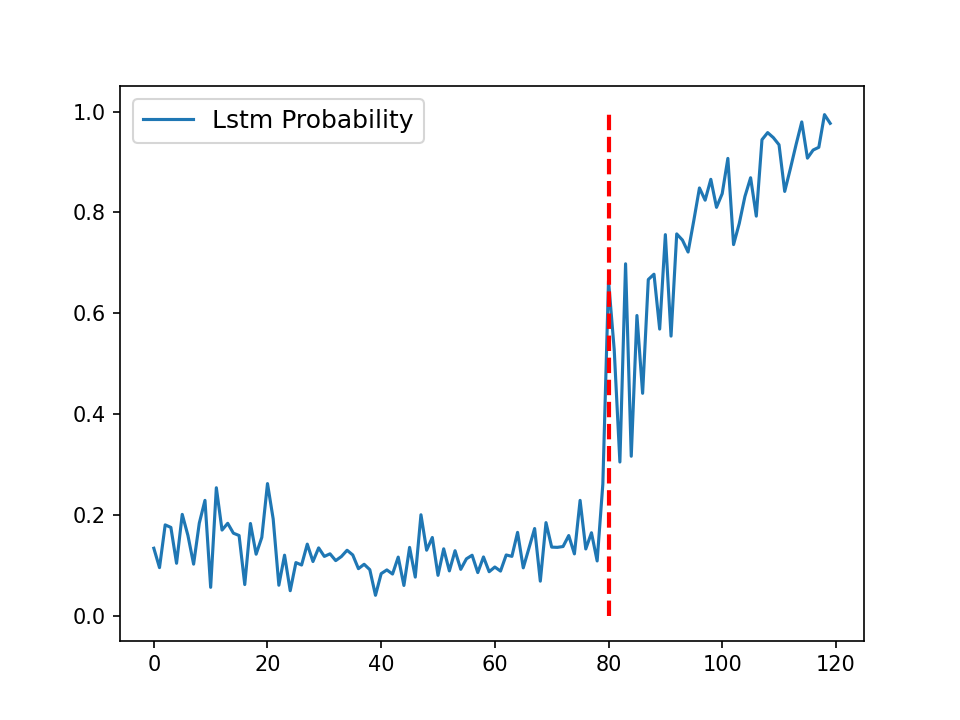}
  }
  \caption{Predicted ictal probabilities (blue lines) and true ictal onset moments (red dashed lines): (a-c) for $|X|=20$ and $|Y|=5$; (d-f) for $|X|=20$ and $|Y|=10$.}
  \label{pre comp}  
\end{figure}

In order to show the effiency of our predictions more intuitively, we compare predicted ictal probabilities with the true ictal moment identified by experienced physicians. Figure~\ref{pre comp} illustrates the results with input duration $|X|=20$ and prediction horizons $|Y|=5$ (Figures~\ref{lstm pre 5},~\ref{renet pre 5},~\ref{meta pre 5}) and $|Y|=10$ (Figures~\ref{lstm pre 10},~\ref{renet pre 10},~\ref{meta pre 10}). The horizontal axis shows 120 consecutive time samples with an interval of 0.5 seconds, and the vertical axis represents the output probability of the model. Solid blue lines show the ictal probabilities learned by different models, and red dashed lines represent the true ictal moment. It is clear that for the LSTM model and ResNet model, the predicted onset time is around 5 seconds late compared to the real onset time. Both models almost always output probability values of $0$ or $1$, with very few moments of output probability between $0$ and $1$. That means that these two models can only learn features of clean data, i.e., they cannot capture the features of noisy data and make a timely prediction. On the contrary, our model has a clear fluctuation that occurs just before and after the onset of the seizure (see the position of the red dashed line in Figures~\ref{meta pre 5} and~\ref{meta pre 10}), which suggests that our model captures the changes in noisy data adequately.

\paragraph{Early warning prediction.}  Early warning of epileptic seizures is of paramount importance for individuals with epilepsy. In addition to the accuracy associated with predicted label generation, our research also places a significant emphasis on evaluating the early warning capacity inherent to the model.



\begin{figure}[!t]
\centering            
  \subfloat[$I_P$ vs. variance, $|X|=20$, $|Y|=5$]
  {
      \includegraphics[width=0.48\textwidth]{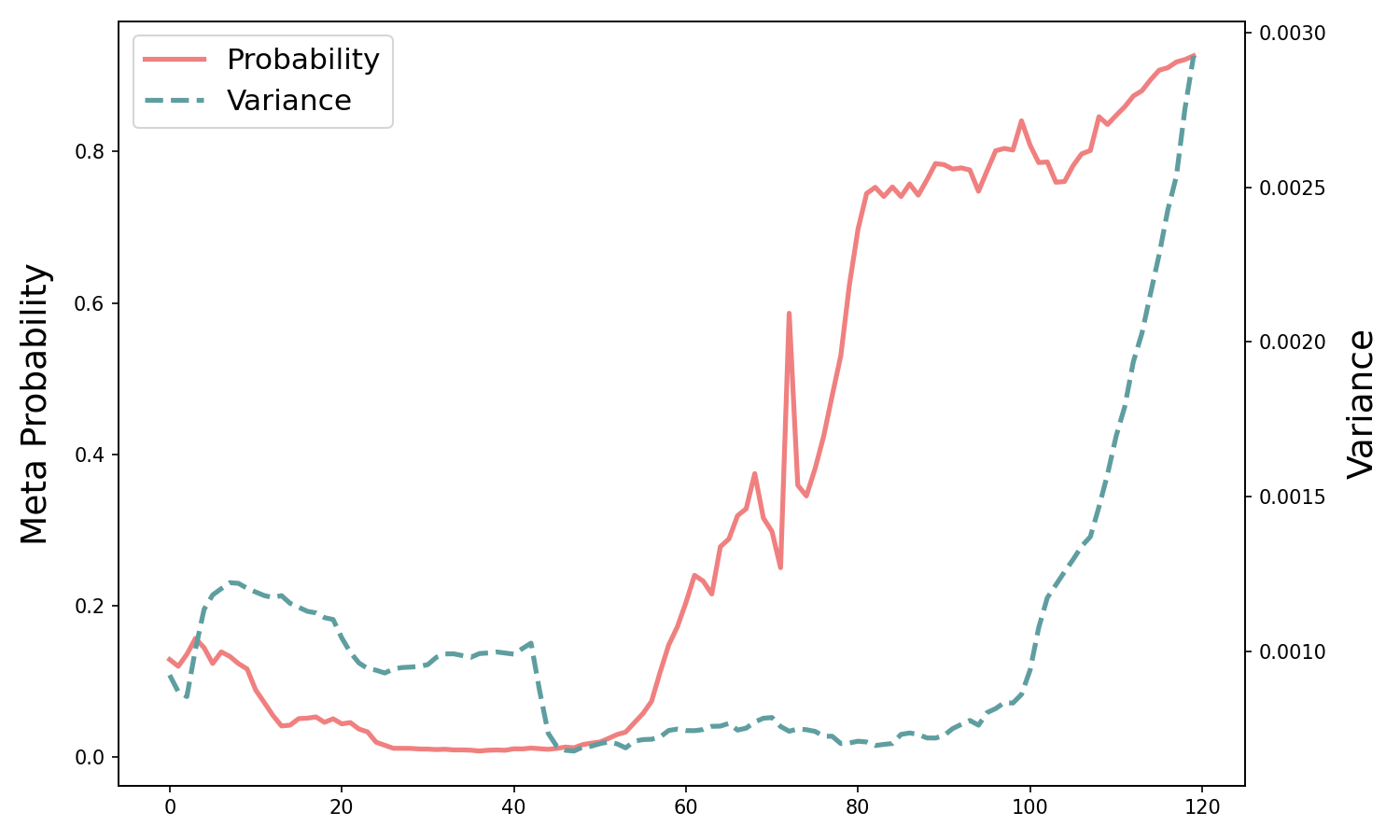}
  }
  \subfloat[$I_P$ vs. variance, $|X|=20$, $|Y|=10$]
  {
    \includegraphics[width=0.48\textwidth]{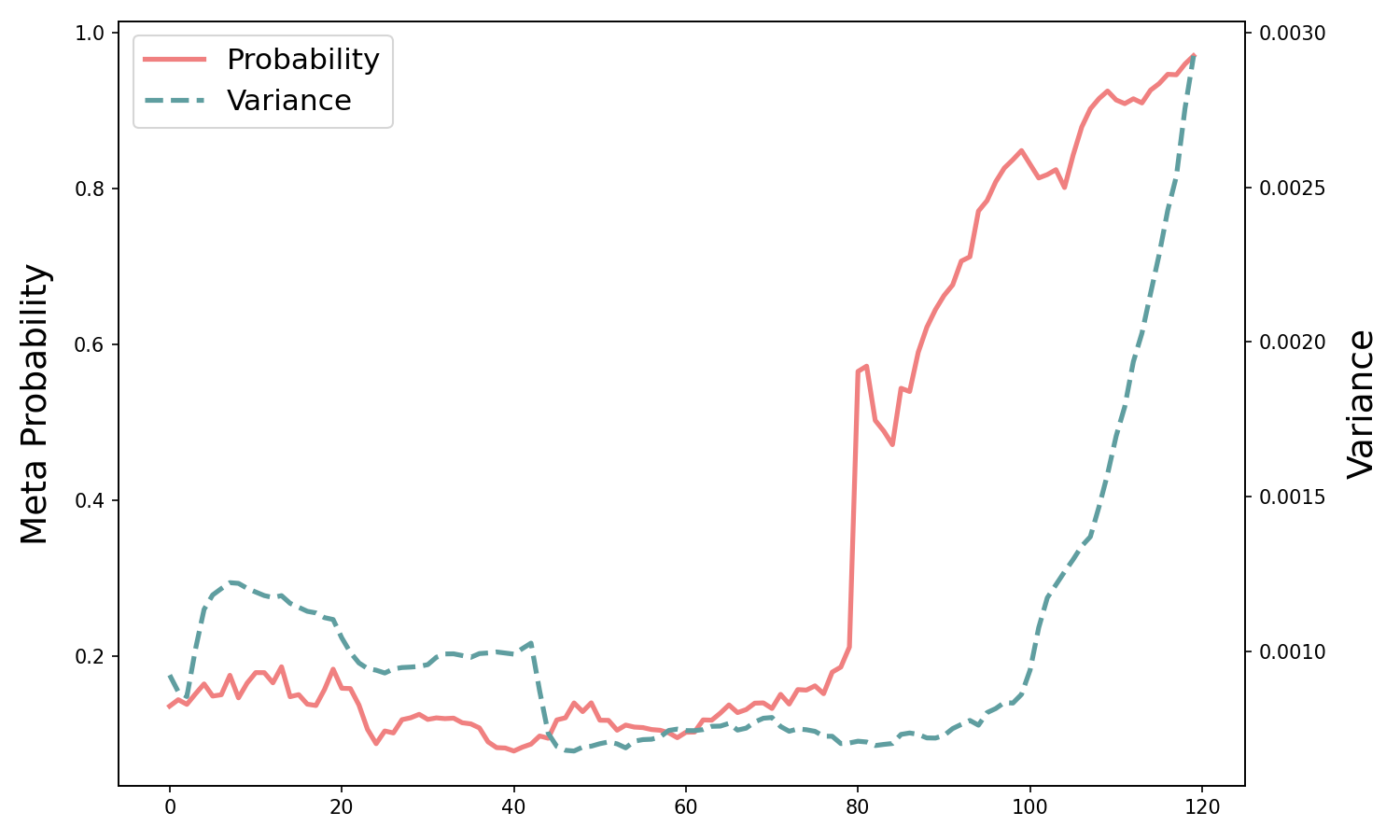}
  }
  
  \caption{A comparison of the proposed indicator $I_{P}$ (red solid line) with the variance indicator (blue dashed line) for EEG data: (a) $|X|=20$, $|Y|=5$; (b) $|X|=20$, $Y=10$.}
  \label{meta_pro_vs_sd}  
\end{figure}

Inspired by \cite{carpenter2006rising}, we compare the tipping signal in terms of our indicator $I_{TP}$ and the variance indicator. Figure~\ref{meta_pro_vs_sd} shows the variance from Figure~\ref{var} averaged over a window of $0.5$ seconds (dashed blue line) 
compared against the predicted probability averaged over five experiments (solid red line), which we call the $I_P$ indicator.
Figure~\ref{meta_pro_vs_sd} illustrates that there is an earlier tipping point in our indicator $I_{P}$ than in the variance indicator on the same data. 
It is clear that for any given reasonable threshold,
our indicator $I_{P}$ always gives an earlier warning than the variance indicator, which suggests that our indicator $I_{P}$ is an effective early warning signal.








\section{Conclusion}
Early warning of epileptic seizures is of paramount significance for individuals afflicted with epilepsy. In this study, we introduce a novel meta-learning framework that can integrate with existing baselines to enhance the predictive capabilities for early ictal signal detection, as evidenced by our results on EEG data obtained from real patients. Our approach amalgamates diverse components (a CNN for the main network and an LSTM for the meta network) to improve prediction accuracy. In addition, we introduce the concept of meta label generation, a method that segregates the original data into clean and noisy segments. This approach leverages the meta label generator to effectively label noisy samples, thereby augmenting the overall data quality. 

Biological, physical, and social systems frequently exhibit qualitative shifts in their dynamics. The development of early warning signals that aim to anticipate these transitions in advance remains a crucial objective for many problem settings, even beyond epileptic seizure detection. We incorporate tipping dynamics as a pivotal feature extracted from the data. Our study demonstrates that the output probability produced by our proposed meta-network is a highly effective early warning indicator. Our experimental findings show a remarkable improvement in prediction accuracy, particularly in scenarios involving long-term predictions. In total, our method offers a robust and innovative approach to the detection of early warning signals, effectively bridging the gap between data-driven models and tipping dynamics.

\section*{Acknowledgements}
We would like to thank Luxuan Yang and Yiheng Ding for their helpful discussions. This work was supported by the National Key Research and Development Program of China (No. 2021ZD0201300), the National Natural Science Foundation of China (No. 12141107), the Fundamental Research Funds for the Central Universities (5003011053) and Dongguan Key Laboratory for Data Science and Intelligent Medicine.

\section*{Data Availability}
The datasets that support the findings of this study are available upon request.

\bibliographystyle{unsrt}
\bibliography{main}

\end{CJK}
\end{document}